\documentclass{article}
\usepackage{spconf,amsmath,graphicx}
\usepackage{amsfonts,amssymb}
\usepackage{booktabs}

\usepackage{algorithm}
\usepackage{algorithmic}
\usepackage{amssymb}
\usepackage{bbding}
\usepackage{amsmath}
\usepackage{multirow}
\usepackage{enumitem}

\usepackage{newfloat}
\usepackage{listings}

\ninept
\usepackage{hyperref}
\hypersetup{
    colorlinks=true
    }

\usepackage{setspace}

\newcommand\blfootnote[1]{%
  \begingroup
  \renewcommand\thefootnote{}\footnote{#1}%
  \addtocounter{footnote}{-1}%
  \endgroup
}

\title{Leveraging Speech PTM, Text LLM, and Emotional TTS \\ for Speech Emotion Recognition}
\name{ 
Ziyang Ma\textsuperscript{1}, 
Wen Wu\textsuperscript{2}, 
Zhisheng Zheng\textsuperscript{1},
Yiwei Guo\textsuperscript{1}, 
Qian Chen\textsuperscript{3}, 
Shiliang Zhang\textsuperscript{3}, 
Xie Chen\textsuperscript{1}$^\ast$
}

\address{
$^1$ MoE Key Lab of Artificial Intelligence, AI Institute,  \\
X-LANCE Lab, Department of Computer Science and Engineering, \\ 
Shanghai Jiao Tong University, Shanghai, China  \\
$^2$Department of Engineering, University of Cambridge, Cambridge, UK \\
$^3$ Speech Lab of DAMO Academy, Alibaba Group, China \\ 
}
   
\begin{document}
\ninept
\maketitle
\begin{abstract}
In this paper, we explored how to boost speech emotion recognition (SER) with the state-of-the-art speech pre-trained model (PTM), data2vec, text generation technique, GPT-4, and speech synthesis technique, Azure TTS. 
First, we investigated the representation ability of different speech self-supervised pre-trained models, and we found that data2vec has a good representation ability on the SER task. 
Second, we employed a powerful large language model (LLM), GPT-4, and emotional text-to-speech (TTS) model, Azure TTS, to generate emotionally congruent text and speech. 
We carefully designed the text prompt and dataset construction, to obtain the synthetic emotional speech data with high quality. 
Third, we studied different ways of data augmentation to promote the SER task with synthetic speech, including random mixing, adversarial training, transfer learning, and curriculum learning. 
Experiments and ablation studies on the IEMOCAP dataset demonstrate the effectiveness of our method, compared with other data augmentation methods, and data augmentation with other synthetic data. 

\end{abstract}

\blfootnote{Corresponding author$^\ast$. }

\section{Introduction}
\label{sec:Introduction}

Speech emotion recognition (SER) is an active research field of intelligent speech processing, both in academia and industry. 
With the development of deep learning methods, the performance of SER has been significantly improved~\cite{latif2021survey}. 
Nonetheless, data scarcity is one of the most important reasons that hinders the further performance improvement of SER. 
Since annotating emotions in speech is time-consuming and subjective, large-scale high-quality labeled data is challenging to obtain. 
Therefore, alleviating the data scarcity problem becomes essential. 

\begin{table}[h]
\centering
\caption{Performance of different SSL pre-trained models on SER task. The setting of the downstream model follows SUPERB~\cite{yang2021superb} to test the representation ability of different upstream models on the IEMOCAP~\cite{busso2008iemocap} dataset. ``LS 960 hr" means LibriSpeech 960 hours, and ``Mix 94k hr" means 94k hours of data including LibriLight, VoxPopuli, and GigaSpeech. }
\vspace{0.3cm}
\label{tab:SSL Models on SUPERB}
\resizebox{1\linewidth}{!}{
\begin{tabular}{lccc}
\hline
\textbf{Model}   & \textbf{\# Params}  & \textbf{Pre-training Corpus}  & \textbf{WA(\%) $\uparrow$} \\
\hline
wav2vec~\cite{schneider2019wav2vec}          & 32.54M              & LS 960 hr                     & 59.79                   \\
vq-wav2vec~\cite{baevski2019vq}       & 34.15M              & LS 960 hr                     & 58.24                   \\ 
wav2vec 2.0~\cite{baevski2020wav2vec}      & 95.04M              & LS 960 hr                     & 63.43                   \\
HuBERT~\cite{hsu2021hubert}           & 94.68M              & LS 960 hr                     & 64.92                   \\
WavLM~\cite{chen2022wavlm}            & 94.70M              & LS 960 hr                     & 65.94                   \\
WavLM+~\cite{chen2022wavlm}           & 94.70M              & Mix 94k hr                    & 67.98                   \\
data2vec~\cite{baevski2022data2vec}         & 93.75M              & LS 960 hr                     & 68.02                   \\ 
\hline
\end{tabular}
}
\end{table}

Using self-supervised learning (SSL) features is an effective way to make up for the lack of labeled data. 
SSL does not require labor-intensive manual annotation, while learning universal representations from a large amount of unlabeled data. 
There are prominent speech self-supervised pre-trained models (PTMs)~\cite{baevski2020wav2vec,hsu2021hubert,chen2022wavlm, baevski2022data2vec,ma2022mt4ssl,yao2022tessp,zhou2022mmspeech,ma2023pushing}, which have been proven to work well across different downstream tasks~\cite{yang2021superb}, such as speech recognition \cite{chen2023reducing}, speaker verification~\cite{chen2022large}, as well as emotion recognition~\cite{wu23_interspeech}. 

In the field of SER, the most commonly used upstream pre-trained models are wav2vec 2.0~\cite{baevski2020wav2vec} and HuBERT~\cite{hsu2021hubert}. 
Some works freeze the upstream model and then use the extracted features to build downstream models~\cite{pepino2021emotion, li2023exploration}. 
Other works finetune the upstream model on emotion recognition corpus and achieve better results~\cite{morais2022speech, chen2023exploring, gat2022speaker}. 
A series of subsequent works comprehensively investigated the SER performance of wav2vec 2.0 and HuBERT model, in the case of no fine-tuning, partial fine-tuning, and entire fine-tuning, either on automatic speech recognition (ASR) corpus or SER corpus~\cite{wang2021fine, wagner2023dawn}. 
Some recent works utilize WavLM~\cite{chen2022wavlm} as the feature extractor, leading to better performance~\cite{ioannides2023towards, chen2023vesper}. 
However, one of the most recent speech PTMs, data2vec~\cite{baevski2022data2vec}, has little presence in the field of SER. 
Following the setting of the SUPERB~\cite{yang2021superb} benchmark, we freeze the data2vec model and conduct a weighted sum of features from different Transformer layers to generate the final representations.
For the downstream model, only linear layers and a pooling layer are employed. 
Table~\ref{tab:SSL Models on SUPERB} compare different speech PTMs for the SER task with weighted accuracy (WA). data2vec achieves better accuracy despite using less pre-training data than other widely used models. 
Therefore, in our experiments, we employ data2vec as the feature extractor, so as to construct a strong baseline. 

\begin{table*}[t]
\centering
\caption{Dataset descriptions and statistics at a glance.}
\vspace{0.3cm}
\label{tab:Dataset}
\resizebox{1\linewidth}{!}{
\begin{tabular}{clc}
\hline
\textbf{Info}   & \textbf{Descriptions}  & \textbf{Statistics}  \\
\hline
\hline
\multicolumn{3}{l}{\textbf{\textit{Text Generation}}} \\
\hline
Narrative Styles          & dialogue, narrative              & 2                      \\
\hline
\multirow{4}*{Scenarios}       & arts, autos and vehicles, business comedy, crime, education,  entertainment, film and animation,     & \multirow{4}*{24}   \\ 
                               & gaming, health and fitness, history, howto and style, kids and family, leisure, music,        & \\
                               & news and politics, nonprofits and activism, people and blogs, pets and animals, & \\
                               & religion and spirituality, science and technology, society and culture, sports, travel and events & \\
\hline
Styles (Emotions)  &angry, cheerful, excited, friendly, hopeful, sad, shouting, terrified, unfriendly, whispering, terrified, neutral & 12 \\
\hline
Max Tokens & 10, 30, 50 & 3 \\
\hline
\hline
\multicolumn{3}{l}{\textbf{\textit{Speech Synthesis}}} \\
\hline
Speakers & 5 females, 4 males & 9 \\
\hline
Styles (Emotions)  &angry, cheerful, excited, friendly, hopeful, sad, shouting, terrified, unfriendly, whispering, terrified, neutral & 12 \\
\hline
\hline
\end{tabular}
}
\end{table*}

Starting with a strong baseline, we attempt to perform data augmentation to further improve the performance. 
Early works conduct data augmentation by modifying or combining the original speech. 
Speed perturbation~\cite{ko2015audio}, SpecAugment~\cite{park2019specaugment} and mixup~\cite{zhang2017mixup} are widely adopted techniques in ASR to enhance the robustness and performance of the systems. 
Research has shown that these tricks can improve the robustness of SER systems~\cite{latif2019direct,lakomkin2018robustness, latif2022multitask, latif2020deep} while the performance gain is limited, since little new emotional information is introduced. 
Adding noise~\cite{lakomkin2018robustness} and applying impulse response~\cite{pappagari2020x} can also help improve the generalization ability of SER systems, while their impact is diminished when SSL models are incorporated for feature extraction~\cite{atmaja2022effects}. 
Recent works leverage generative models to create additional training data, so as to improve the performance of SER. 
Methods of adding synthetic data are not limited by front-end features, which cooperate well with SSL features. 
Popular methods include using GAN-based models~\cite{bao2019cyclegan,rizos2020stargan} and diffusion-based models~\cite{malik2023preliminary}. 
These methods use the original training data to train generative models, thereby generating synthetic data consistent with the original distribution. 
Little has been investigated on how to incorporate generated speech with the original speech during training in cases of cross-corpus and cross-domain. 
Recent work~\cite{zhang2023refashioning} investigates the capabilities of generative LLM in the field of affective computing. However, their work lies in textual emotions and sentiments, and there is still a blank in the field of speech emotion recognition. 

Our goal is to leverage state-of-the-art text generation and speech synthesis techniques to synthesize high-quality speech, and effectively fuse it with the original speech to enhance SER performance. 
For text generation, we use GPT-4~\cite{openai2023gpt4}, one of the best-performing large language mode (LLM) to synthesize text with emotional expressions. 
Careful prompt design and delicate data engineering are performed to ensure the quality of text obtained. 
We utilize the generated text to synthesize speech using Azure text-to-speech (TTS)~\footnote{https://learn.microsoft.com/en-us/azure/ai-services/speech-service/text-to-speech} with emotionally congruent expressions, which guarantees acoustic and semantic consistency. 
For example, a person would almost never say a sad text in a happy tone. 
We explored different ways of data augmentation, including random mixing, adversarial training, transfer learning, and curriculum learning, to make synthetic data helpful. 
We also test our method with EmoDiff~\cite{guo2023emodiff}, a state-of-the-art diffusion-based specialist model for emotional TTS. 
Experiments and ablation studies on the IEMOCAP~\cite{busso2008iemocap} dataset demonstrate the effectiveness of the proposed method.

\section{Proposed Approach}
\label{sec:Method}

In this section, we first introduce how to synthesize high-quality emotional speech data, and then present how to use the synthetic data to improve the performance of speech emotion recognition. 
\subsection{Data Synthesis Methods}

\subsubsection{Dataset Configuration}
There are two steps to synthesize semantically consistent emotional speech: 1) Generate emotional text data. 2) Synthesize emotional speech data rich in corresponding emotions using the generated text. 
In order to accomplish the above goals, we design a setting to construct the synthetic dataset, as shown in Table~\ref{tab:Dataset}, where we list the elements involved in text generation and speech synthesis. 

When configuring text generation, we mainly consider 4 kinds of information:
\begin{enumerate}
    \item \textbf{Narrative Styles.} In the real world, human speech comes either from a conversation scene (dialogue) or from reading some text (narrative). 
    \item \textbf{Scenarios.} Referring to the classification method of GigaSpeech~\cite{chen2021gigaspeech}, we classify the scenarios into 24 classes. 
    \item \textbf{Styles (Emotions).} According to the predefined styles of Azure Emotional TTS, We choose 11 of them, along with 1 additional ``neutral" emotion. The emotions can be taken as a subset of the styles. 
    \item \textbf{Max Tokens.} In order to make the subsequent synthesized speech diverse in length distribution, we make the generated text with different ranges of token numbers. We use max token lengths to represent short, medium, and long sentences. 
\end{enumerate}

When configuring speech synthesis, we mainly consider 2 kinds of information:
\begin{enumerate}
    \item \textbf{Speakers.} To train a SER model, different speakers are usually required to increase the robustness of the model. Here we employ 5 female timbres and 4 male timbres. 
    \item \textbf{Styles (Emotions).} We generate emotionally congruent text and speech, therefore the number of styles (emotions) remains consistent with the text, which is 12 classes. 
\end{enumerate}

\begin{algorithm*}[t]
\caption{Prompt Engineering for Emotional Text Generation}
\label{algo:prompt}
\begin{algorithmic}[1]
\REQUIRE{Input: lists of Narrative\_Styles $\mathbf{N}$, Scenarios $\mathbf{S}$, Emotions $\mathbf{E}$, Max\_Tokens $\mathbf{M}$}
\ENSURE{Output: lists of generated text $\mathbf{U}$ from GPT-4}
\STATE{Set the system role: ``You are a helpful assistant with human emotions and talking styles."}
\STATE{Group tuples $\mathbf{T}$ which meets the requirements that $\textit{narrative\_style} \in \mathbf{N}$, $\textit{scenario} \in \mathbf{S}$, $\textit{emotion} \in \mathbf{E}$, $\textit{max\_token} \in \mathbf{M}$, and $(\textit{narrative\_style}, \textit{scenario}, \textit{emotion}, \textit{max\_token}) \in \mathbf{T}$}
\FOR{each tuple $\mathbf{t} \in \mathbf{T}$}
        \STATE{Map $\textit{max\_token}$ to text description that \textit{length2str}:\{10: short, 30: middle, 50: long\} }
        \IF{\textit{narrative\_style} $=$ `` dialogue "}
        \STATE{\textit{prompt} $=$ `` In the context of \textit{scenario}, say something in first-person or second-person that expresses your feeling, or using the speaking style of \textit{emotion}, as if you are talking to somebody. Do not write any explanations and just answer the question. What you say should be \textit{length2str}(\textit{max\_token}) length with no more than \textit{max\_token} words. "}
        \ELSIF{$\textit{narrative\_style} = $ `` narrative "}
        \STATE{\textit{prompt} $=$ `` In the context of \textit{scenario}, describe a third-person scene that conveys the emotion, or using the speaking style of \textit{emotion}. Do not write any explanations and just answer the question. What you say should be \textit{length2str}(\textit{max\_token}) length with no more than \textit{max\_token} words. "}
        \ENDIF
        \STATE{Sample a batch of generated text $\mathbf{u}$ with \textit{prompt} and add $\mathbf{u}$ to $\mathbf{U}$}
\ENDFOR
\STATE{Perform data engineering on the generated text $\mathbf{U}$ to ensure the quality of generation}
\RETURN{$\mathbf{U}$ }
\end{algorithmic}
\end{algorithm*}

\subsubsection{Text Generation}
To synthesize high-quality text with emotional expressions, we use one of the most powerful LLM available, GPT-4~\cite{openai2023gpt4}, for this task. 
A key point to the successful use of GPT-4 is to design a good prompt. 
Algorithm~\ref{algo:prompt} presents our prompt for emotion text generation. 
Given lists of narrative styles, scenarios, emotions, and max tokens, we group tuples from them. 
For each tuple, we choose the corresponding prompt according to the narrative style and fill in the rest content into the prompt. 
We then sample a batch of generated text from GPT-4 and add them to the output pool. 
Finally, we perform data cleaning on the generated text to guarantee the generated text is exactly what we need for the subsequent speech synthesis. 

\subsubsection{Speech Synthesis}
Once the generated text is obtained, we synthesize speech that is semantically consistent with the text. 
We implement this step by using Azure Emotional TTS with Speech Synthesis Markup Language (SSML)~\cite{taylor1997ssml}. 
The total length of synthesis speech is more than 500 hours, which is far more than the size of the IEMOCAP dataset. 
We select a subset to conduct the experiments, which will be detailed in Section~\ref{sec:Synthetic Datasets}.

\subsection{Model Configuration and Training}
For the upstream model, we use a data2vec model with parameters frozen. 
In other words, we use a pre-trained data2vec model as the feature extractor to extract speech representations of 768 dimensions. 
Representations using the last layer or multi-layer weighted with learnable parameters are both considered. 
The learnable parameters are initialized with an average of layer numbers, and optimized during the training process of the downstream model.

For the downstream model, following the common practice of SUPERB~\cite{yang2021superb}, we use a linear layer to map 768 dimensions to 128 dimensions, followed by a ReLU activation function and an average pooling layer, and finally use a linear classification layer to obtain the probability distribution of each emotion. 
The cross-entropy loss is employed for optimizing the downstream model. 

For the training procedure, we use the AdamW optimizer and set the learning rate to 1e-3, and the weight decay to 2e-3. 
We train the model on a single Nvidia RTX 3090 GPU with a batch size of 128 for 50 epochs. 

For the inference procedure, we apply widely employed evaluation metrics, weighted accuracy (WA) and unweighted accuracy (UA), to evaluate the performance of speech emotion recognition. 
WA corresponds to the overall accuracy while UA corresponds to the average class-wise accuracy.

For data augmentation with synthetic speech, simply mixing the synthetic with the original does not improve the performance. 
Here we explore different data augmentation methods, say, adversarial training, transfer learning, and curriculum learning. 
The details and results of different ways of data augmentation are clarified in Section~\ref{sec:Results with Different Augmentation Methods}. 

\vspace{-0.3cm}
\section{Experiments}
\label{sec:Experiments}

\subsection{Datasets Details}
\subsubsection{Real Datasets}
\label{sec:Real Datasets}
The benchmark dataset IEMOCAP~\cite{busso2008iemocap} is used for SER. It consists of 5 dyadic conversational sessions performed by 10 professional actors with a session being a conversation between two exclusive speakers. Each utterance was annotated by three human annotators and the ground-truth labels were determined by majority vote. Following prior work~\cite{yang2021superb}, four emotion classes are used: happy (merged with excited), sad, angry, and neutral, which give 5531 utterances. Leave-one-session-out 5-fold cross-validation setup is used. In each fold, one session is held out for testing, while the remaining four sessions are utilized for training and validation, split in an 8:2 ratio. The average results cross folds are reported. 

\subsubsection{Synthetic Datasets}
\label{sec:Synthetic Datasets}
By observing the IEMOCAP dataset, we found that almost all of the speech in IEMOCAP is short speech. 
We choose the speech synthesized by sentences with a max token of 10. 
Besides, emotions in synthetic speech include happy sad, angry, and neutral, and the narrative style is dialogue, both of which keep consistent with the IEMOCAP dataset. 
This results in a subset of 6473 sentences, which outcomes about 8.8 hours of speech. 
Due to the difference in distribution between the augmented data and the original data, when training with synthetic speech, we select the checkpoint of the last epoch instead of the best checkpoint on the validation set.

\subsection{Results with Different Augmentation Methods}
\label{sec:Results with Different Augmentation Methods}
Due to the difference in the distribution of synthetic data and original data, mixing directly does not improve the performance of the model. 
As shown in Table~\ref{tab:Results augmentation methods}, we explore several different data augmentation methods, where experiments are based on the features extracted from the last layer of the data2vec model. 

\textbf{Random mixing. } 
We just add the synthetic speech to the original speech and shuffle them. Compared with the baseline, the performance of the SER task does not improve due to the difference with respect to the distribution. 

\textbf{Adversarial training. }
In the case of adversarial training, the model contains three sub-networks, namely feature fuser, SER classifier, and domain classifier~\cite{ganin2016domain}. 
The feature fuser conducts dimension transformation on multi-layer or single-layer features and pooling operation. 
The SER classifier employs cross-entropy loss for emotion classification, while the domain classifier employs binary cross-entropy loss for domain classification. 
Each batch is optimized in three steps. 
The first step is to calculate the loss of emotion classification, and then optimize the feature fuser and the SER classifier simultaneously. 
The second step is to calculate the loss of domain classification, and then only optimize the domain classifier, with the gradient of the feature fuser detached. 
The third step is to pass the domain classification loss back to the feature fuser, perform gradient reversal, and optimize the feature fuser. 
Through adversarial training, data with different distributions will be adjusted to a similar distribution within the feature fuser. 
However, through this training criterion, the performance of SER is not greatly improved. 
The reason may be that this method is designed to adapt the new domain to the original domain, while the effect on improving the performance of the original domain is limited. 

\textbf{Transfer learning. }
Transfer learning is a training method that transfers learned knowledge to target data. 
Under this paradigm, we first train the SER model on synthetic speech, then reduce the learning rate, and then migrate to the IEMOCAP dataset for training. 
Through transfer learning, the performance of the model is effectively improved. 

\textbf{Curriculum learning. }
Curriculum learning is a training strategy that organizes the learning process by gradually increasing the complexity of training samples. 
In this case, we sort the augmented data from short to long, and gradually add it to the original data every fixed number of epochs during the training phase. 
The model has great performance gain on the IEMOCAP dataset in this way. 
\vspace{-0.3cm}

\begin{table}[htbp]
\centering
\caption{Results with different augmentation methods compared with the baseline. WA and UA are reported here. }
\label{tab:Results augmentation methods}
\vspace{0.3cm}
\begin{tabular}{ccc}
\hline
\textbf{Method}           & \textbf{WA(\%) $\uparrow$}  & \textbf{UA (\%)$\uparrow$}   \\
\hline
baseline                  & 64.52                   & 68.22                     \\
random mixing             & 64.24                 & 68.06                           \\ 
adversarial training      & 64.88                  & 68.85                               \\
transfer learning         & 68.20                   & 70.72                             \\
curriculum learning       & 68.57                   & 70.86      \\
\hline
\end{tabular}
\end{table}

\vspace{-0.6cm}
\subsection{Results with Different Amounts of Augmentation Data}
\label{sec:Results with Different Amounts of Augmentation Data}
\begin{figure}[htbp]
    \centering
    \includegraphics[width=1.0\linewidth]{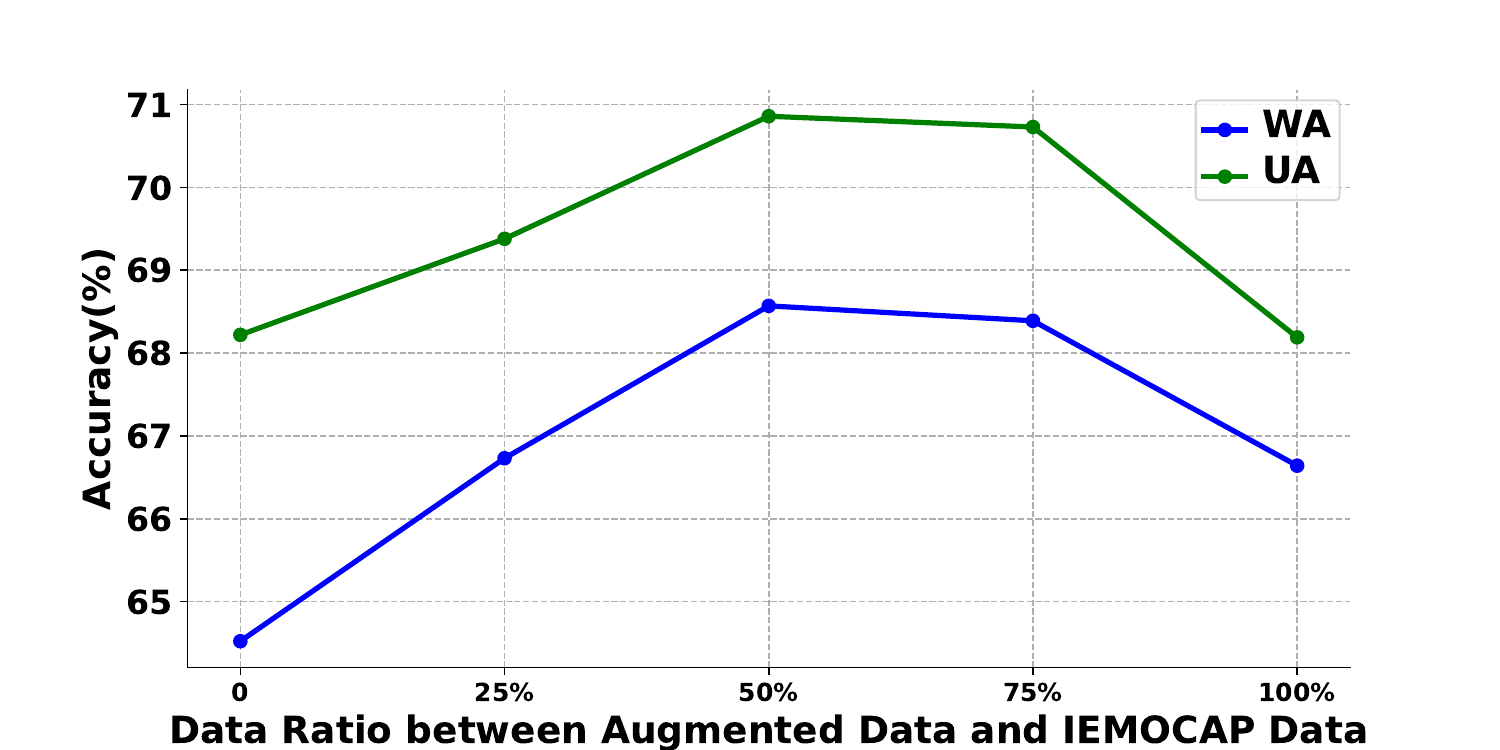}
    \caption{Results with different amounts of augmentation data. Both WA and UA are drawn out here. }
    \label{fig:Results Amounts}
\end{figure}

We found that adding more augmented data is not always more profitable. 
Based on the optimal configuration from Section~\ref{sec:Results with Different Augmentation Methods}, we control the total amount of synthetic data added. 
As shown in Figure~\ref{fig:Results Amounts}, both weighted accuracy and unweighted accuracy become better as the amount of synthetic data increases. 
The model achieves best performance when the amount of synthetic data added is half of the original data. 
When the synthetic data continues to be added, the performance of the model begins to decline. 
One possible reason is that adding too much synthetic data will allow the distribution of synthetic data to dominate the training progress. 
\vspace{-0.3cm}

\vspace{-0.5cm}
\subsection{Results on the SER Task}
With a suitable augmentation method and a suitable amount of data, we train the downstream model with our method on the SER task. 
More specifically, we use the training strategy of curriculum learning and set the data ratio to 1:2 between synthetic data to real data. 
Extracted features employing the last layer or multi-layer weighted from the pre-trained data2vec model are both considered. 
We compare our method with the baseline without synthetic data. 
We also compare with models augmented with synthetic data generated from EmoDiff~\cite{guo2023emodiff}, the most recent emotional TTS utilizing diffusion technique, with the same emotional text. 
As shown in Table~\ref{tab:Results}, our method improves the SER performance on both weighted accuracy and unweighted accuracy. 
Experiments with EmoDiff also improve compared to the baseline, indicating that our method can be transferred to other emotional TTS. 

\vspace{-0.3cm}
\begin{table}[htbp]
\centering
\caption{Results on SER task compared with baselines. Representations using the last layer or multi-layer weighted with learnable parameters are both considered. WA and UA are reported here. }
\vspace{0.3cm}
\label{tab:Results}
\begin{tabular}{llcc}
\hline
\textbf{Representations} &\textbf{Training Data}    & \textbf{WA $\uparrow$}  & \textbf{UA $\uparrow$}   \\
\hline
\multirow{3}*{last layer} &IEMOCAP   & 64.52                   & 68.22               \\
&IEMOCAP + EmoDiff & 68.02                  & 69.67           \\ 
&\textbf{IEMOCAP + ours} &  \textbf{68.57}               &   \textbf{70.86}           \\ 
\hline
\multirow{3}*{multiple layers} &IEMOCAP   & 68.02                   & 69.95               \\
&IEMOCAP + EmoDiff &  68.39                 &  71.06          \\ 
&\textbf{IEMOCAP + ours}  & \textbf{68.85}                  & \textbf{71.89}           \\ 
\hline
\end{tabular}
\end{table}

\vspace{-0.6cm}
\section{Conclusion}
\label{sec:conclusion}
\vspace{-0.1cm}
In this paper, we aim to leverage state-of-the-art techniques for augmenting speech emotion recognition. 
We utilize an awesome large language model contemporarily, GPT-4, to generate text with emotional expressions, and Azure Emotional TTS, to synthesize speech with emotional consistency. 
We also leverage a powerful speech-based pre-trained model, data2vec, as the feature extractor, to extract features of synthetic data and real data. 
Besides, effective augmentation strategies and suitable augmented data volume are explored. 
Experiments on the IEMOCAP dataset demonstrate the effectiveness of our method. 
In our experiments, only a small subset of our synthetic data is used. In the future, we will explore how to use large-scale high-quality synthetic data to enhance cross-corpus, cross-domain, and cross-language situations. 

\vfill\pagebreak
\begin{spacing}{0.01}
\footnotesize
\bibliographystyle{IEEEbib}
\bibliography{strings,refs}
\end{spacing}

\end{document}